\documentclass[sn-mathphys-num]{sn-jnl}% Math and Physical Sciences Numbered Reference Style 
\usepackage{graphicx}%
\usepackage{multirow}%
\usepackage{amsmath,amssymb,amsfonts}%
\usepackage{amsthm}%
\usepackage{mathrsfs}%
\usepackage[title]{appendix}%
\usepackage{xcolor}%
\usepackage{textcomp}%
\usepackage{manyfoot}%
\usepackage{booktabs}%
\usepackage{algorithm}%
\usepackage{algorithmicx}%
\usepackage{algpseudocode}%
\usepackage{listings}%
\theoremstyle{thmstyleone}%
\theoremstyle{thmstyletwo}%

\theoremstyle{thmstylethree}%

\begin{document}

\title[Real-Time Anomaly Detection with Synthetic Anomaly Monitoring (SAM)]{Real-Time Anomaly Detection with Synthetic Anomaly Monitoring (SAM)}

%%=============================================================%%
%% GivenName	-> \fnm{Joergen W.}
%% Particle	-> \spfx{van der} -> surname prefix
%% FamilyName	-> \sur{Ploeg}
%% Suffix	-> \sfx{IV}
%% \author*[1,2]{\fnm{Joergen W.} \spfx{van der} \sur{Ploeg} 
%%  \sfx{IV}}\email{iauthor@gmail.com}
%%=============================================================%%

\author*[1]{\fnm{Emanuele} \sur{Luzio}}\email{emanuele.luzio@mercadolibre.com}

\author[2,3]{\fnm{Moacir} \sur{Antonelli Ponti}}\email{moacir.ponti@mercadolivre.com}
\equalcont{These authors contributed equally to this work.}

\affil*[1]{\orgdiv{MeliMinds}, \orgname{Mercadolibre Inc}, \orgaddress{\street{Luis Bonavita, 1266}, \city{Montevideo}, \postcode{11300}, \state{Montevideo}, \country{Uruguay}}}

\affil[2]{\orgdiv{MeliMinds}, \orgname{Mercadolibre Inc.}, \orgaddress{\street{Av. das Na\c{c}\~{o}es unidas 3003}, \city{Osasco}, \postcode{06233-903}, \state{S\~ao Paulo}, \country{Brazil}}}

\affil[3]{\orgdiv{ICMC - Instituto de Ci\^{e}ncias Matem\'{a}ticas e de Computa\c{c}\~{a}o}, \orgname{Universidade de S\~ao Paulo}, \orgaddress{\street{Av. Trabalhador S\~aocarlense, 400}, \city{S\~{a}o Carlos}, \postcode{ 13566-590}, \state{S\~{a}o Paulo}, \country{Brazil}}}

%%==================================%%
%% Sample for unstructured abstract %%
%%==================================%%

\abstract{
Anomaly detection is essential for identifying rare and significant events across diverse domains such as finance, cybersecurity, and network monitoring. This paper presents Synthetic Anomaly Monitoring (SAM), an innovative approach that applies synthetic control methods from causal inference to improve both the accuracy and interpretability of anomaly detection processes. By modeling normal behavior through the treatment of each feature as a control unit, SAM identifies anomalies as deviations within this causal framework. We conducted extensive experiments comparing SAM with established benchmark models, including Isolation Forest, Local Outlier Factor (LOF), k-Nearest Neighbors (kNN), and One-Class Support Vector Machine (SVM), across five diverse datasets, including Credit Card Fraud, HTTP Dataset CSIC 2010, and KDD Cup 1999, among others. Our results demonstrate that SAM consistently delivers robust performance, highlighting its potential as a powerful tool for real-time anomaly detection in dynamic and complex environments.}


%%\pacs[JEL Classification]{D8, H51}

%%\pacs[MSC Classification]{35A01, 65L10, 65L12, 65L20, 65L70}

\maketitle

\section{Introduction}\label{sec1}

Anomaly detection is the process of identifying data points, events, or patterns that deviate significantly from the expected behavior within a dataset. It requires a definition of normal (not in the sense of the Gaussian distribution, but as not-anomalous) or expected behavior \cite{Hawkins_1980}. The most common definition for anomalies include the following characteristics~\cite{goldstein2016comparative}:
\begin{enumerate}
    \item Anomalies are different from the norm with respect to their features and
    \item They are rare in a dataset compared to normal instances.
\end{enumerate}

The challenge of defining what is a norm with respect to the data. This is often expressed as the difficulty in modeling the probability distribution \( p(X|\theta) \) of the normal data, where \( X \) represents the data and \( \theta \) denotes the parameters of the model. The lack of a clear boundary complicates the task of identifying deviations, as instances that are near the boundary may be misclassified \cite{ Schmidl_2022}.

Detecting anomalies was first applied to dataset cleaning~\cite{goldstein2016comparative}, however more recently it became important for other applications such as the creation of alerts for deviations during monitoring, to mine for particular interesting events or suspicious data records. 
Therefore, it plays a relevant role in various applications, ranging from detecting financial fraud to ensuring system health and security~\cite{10.1145/1541880.1541882}. Despite its critical importance, effectively identifying anomalies remains a challenging task, given the demands for real-time processing, the need for models to generalize to previously unseen data distributions~\cite{Schmidl_2022}, but also requirements of real-world applications, such as explaining the anomalous behavior.

Anomalies can appear in different flavors, such as point, contextual and collective \cite{calikus2022wisdom}. In this paper, we deal with point anomalies, as an individual data point is inspected to be with respect to the rest of data. In terms of methods, according to \cite{10.1145/1541880.1541882}, we can categorize into statistical methods, machine learning methods, data mining methods, information theory and spectral methods.

% Moacir: I introduced the method in the next paragraphs, since it is paramont to tell the contribution as soon as possible

This paper proposes a different type of method based on synthetic control, which is a widely used tool in policy evaluation, including assessments of economic interventions, health policy implementations, and environmental regulations. Their application has extended into various fields, demonstrating their versatility and effectiveness in causal inference from observational data. 
Synthetic control methods enhance causal inference in contexts where traditional regression approaches may fall short, providing a transparent and data-driven way to estimate treatment effects in complex observational settings. 

Our proposed method, called SAM (Synthetic Anomaly Monitoring), models each feature as a control unit and frames anomalies as causal ``treatments/interventions'' on such feature.  It  works by computing counterfactuals for each feature as well as an error for the observed feature value w.r.t. the counterfactual.
We fit a model for the generation of features' counterfactual from which we compute residual errors, and then an anomaly score for a data point. Therefore, our method allows not only to detect anomalies but also to explain each features' contribution to the anomalous behavior, which is a quality of few anomaly detection methods have~\cite{li2023explainable}. SAM contributions are as follows:
\begin{itemize}
    \item frames anomalies as ``treatments'' within a causal inference context, a novel approach that allows for causal reasoning in anomaly detection, enhancing interpretability;
    \item unlike methods that operate on isolated instances (e.g., LOF, KNN), SAM explicitly models relationships between features. This aspect can lead to a more sophisticated understanding of normal behavior patterns within data;
    \item can be efficiently computed in real-time scenarios, potentially outperforming more computationally intensive methods in high-throughput environments;
    \item by constructing individual counterfactuals for each feature, SAM provides interpretable results, offering insights into which features are contributing to an anomaly and to what extent, unlike more opaque models;
    \item SAM allows for online learning, since it does not depend on re-training the whole model for each new instance. 
\end{itemize}

As far as we know this is the first anomaly detection approach using the synthetic control framework. In the following, we are going to briefly describe anomaly detection approaches as well as the challenges involving anomaly detection that justify the investigation of a synthetic control-based method. Most of the content and discussion of previous studies and the field can be found in more detail in \cite{10.1145/1541880.1541882}, \cite{goldstein2016comparative}, \cite{steinbuss2021benchmarking} and \cite{bouman2024unsupervised}.

\section{Anomaly detection (AD)  approaches and Synthetic Control (SC)}

Machine learning methods for anomaly detection involve training a model \( f: \mathbb{R}^n \rightarrow \{0,1\} \) on labeled data \( \{(x_i, y_i)\}_{i=1}^N \), where \( x_i \) represents the feature vector and \( y_i \) indicates whether \( x_i \) is normal (\( y_i = 0 \)) or anomalous (\( y_i = 1 \)). The objective is to minimize a loss function \( L(f(x), y) \) over the training data, such as:
\[
\min_f \sum_{i=1}^N L(f(x_i), y_i),
\]
where \( L \) could be the cross-entropy loss for classification problems. Since this would require having labels from both normal and anomalous classes, often not available or scarce, methods were developed to learn with few anomalous data points~\cite{costa2013partially}, having only normal data e.g. one-class SVM~\cite{scholkopf2001estimating,tax2004support}

Another type of machine learning method involve isolating anomalies from nominal data requiring no annotation. The most relevant method in this context is Isolation Forest~\cite{liu2008isolation} that isolates every single instance using decision trees, and define a anomaly score based on the average path lengths.

Data mining approaches are usually clustering or nearest-neighbor techniques~\cite{ramaswamy2000efficient}, that operate without the need for labeled data. Those are often used as baseline methods. In clustering-based methods, data points are grouped into clusters \( \{C_k\}_{k=1}^K \) based on a similarity measure \( d(x_i, x_j) \). An instance \( x \) is considered an anomaly if it lies far from the nearest cluster, i.e.:
\[
\min_k d(x, C_k) > \tau,
\]
where \( \tau \) is a threshold that defines the maximum allowable distance from the nearest cluster. For nearest-neighbor methods, the anomaly score for a data point \( x \) is given by the distance to its \( k \)-th nearest neighbor \( x_{(k)} \):
\[
\text{score}(x) = d(x, x_{(k)}).
\]

%\subsection{Challenges in Anomaly Detection}

In parallel, synthetic control methods have emerged as a relevant statistical tool in econometrics, particularly for evaluating the impact of policy interventions and treatments in observational studies \cite{10.1257/jel.20191450}. These methods involve constructing a synthetic version of the treatment unit using a weighted combination of control units that were not affected by the intervention. It allows for the estimation of the counterfactual—what would have happened to the treatment unit had the intervention not taken place. This is done by comparing the treated unit to a weighted combination of control units that approximate the characteristics of the treated unit before the intervention. 

Formally, we have:
\[
Y_{it} = \alpha_i + \theta_t + \lambda_t \mathbf{Z}_i + \beta_t \mathbf{W}_i + \epsilon_{it},
\]
where the variables have the following meaning:
\begin{itemize}
    \item \(Y_{it}\): Outcome of interest for unit \(i\) at time \(t\).
    \item \(\alpha_i\): Unit-specific fixed effects.
    \item \(\theta_t\): Time-specific fixed effects.
    \item \(\mathbf{Z}_i\): Vector of observed covariates for unit \(i\).
    \item \(\mathbf{W}_i\): Unobserved variables affecting the outcome.
    \item \(\lambda_t, \beta_t\): Coefficients capturing the impact of observed and unobserved variables over time.
    \item \(\epsilon_{it}\): Error term.
\end{itemize}

By constructing a synthetic control that closely matches the treated unit's observed characteristics (\(\mathbf{Z}_i\)), the method aims to account for both observed and unobserved confounding factors, assuming that unobserved variables (\(\mathbf{W}_i\)) are adequately balanced between treated and control units. To assess and improve the robustness of the estimates against biases from unobserved variables, researchers employ techniques such as placebo tests and sensitivity analyses \cite{10.1198/jasa.2009.ap08746}.

\subsection{Integration of SC into AD}

In anomaly detection, synthetic control methods offer an interesting framework for modeling normal behavior and identifying deviations. By treating anomalies as interventions or treatments, the SAM approach adapts synthetic control principles for real-time data monitoring and analysis, providing a robust mechanism for detecting significant changes in data patterns.

This application extends synthetic control methods beyond their traditional uses in econometrics and policy analysis, demonstrating their versatility in data science and anomaly detection.

We propose \textbf{Synthetic Anomaly Monitoring (SAM)}, leveraging causal inference to model and detect anomalies in real-time data streams. By treating each data feature as a control unit and framing anomalies as causal 'treatments', SAM enhances model adaptability by learning relations between features, ensuring real-time processing by it's linear nature that allows it to be written as a dot product, and improves interpretability since counterfactuals are computed for each individual feature.

The subsequent sections will detail the methodology behind SAM, explore its implementation, and provide comparative analyses with existing benchmark models to demonstrate its efficacy across various domains and challenges.

\section{Conceptual and Mathematical Framework of SAM}

SAM leverages synthetic control methods from causal inference by framing anomalies as treatments within a causal context. Let \( X_i \) represent our feature of interest, which can be expressed as:
\begin{equation}
X_i = S_i + \theta_i(t),
\end{equation}
where \( S_i \) is the ground truth of the feature, and \( \theta_i(t) \) represents the anomalous (or treatment) signal. The anomalous signal is a time-dependent random value. By monitoring the difference between the observed feature \( X_i \) and the ground truth \( S_i \), we can identify deviations from the true value.

The key question is how do we determine the true value \( S_i \). Our hypothesis is that we can infer \( S_i \) using the other features \( X_{j \neq i} \), as follows:
\begin{equation}
S_i = \sum_{j \neq i} \beta_{ij} X_j + \alpha_i.
\end{equation}

The problem involves training a model such that \( \beta_{ii} = 0 \) for all \( i \). There are two possible formulations. One more general approach to solve it involves fitting a single model with the following loss function:
\begin{equation}
\mathcal{L} = \sum \left( X_i - \alpha_i - \sum_{j \neq i} \beta_{ij} X_j \right)^2 + \lambda_i L^n(\beta_{ii}).
\end{equation}

In this formulation:
\begin{itemize}
    \item \( \beta_{ij} \) are the coefficients quantifying the relationship between feature \( X_j \) and feature \( X_i \).
    \item \( \alpha_i \) is the intercept term for feature \( X_i \).
    \item \( \lambda_i \) is the regularization coefficient, with \( \lambda_i \gg 1 \). \( L^n \) denotes the type of regularization (e.g., L1, L2, or Elastic Net), which forces the diagonal elements \( \beta_{ii} \rightarrow 0 \).
\end{itemize}
Using L1 (Lasso), L2 (Ridge), or Elastic Net regularization on the diagonal elements of the \( \beta \)-matrix drives these diagonal elements \( \beta_{ii} \) toward zero.

The second, and simpler approach, is to fit separate linear models for each feature. For simplicity, this was the approach used in this paper.

\subsection{Key Characteristics of the SAM Model}

The SAM approach focuses on capturing relationships between different features rather than learning a partition of the latent space. In this regard, SAM shares certain similarities with autoencoders~\cite{pereira2020reviewing, resende2022robust}, but instead it is specifically designed to prevent any feature from contributing to the prediction of itself. 

The effectiveness of synthetic control methods (SCM) relies heavily on specific methodological considerations, particularly regarding data requirements and the potential for bias in estimates to overcome modeling challenges \cite{10.1257/jel.20191450}.
The reliability of the Synthetic Control Method (SCM) largely depends on the availability and quality of data. It specifically requires a well-defined donor pool of untreated units and a sufficient number of pre-intervention periods to accurately construct a counterfactual scenario. We assume that with ample data, SAM can effectively select the optimal donor pool. However, incorporating RANSAC in the fitting step can offer a possible improvement.

RANSAC (Random Sample Consensus) \cite{Fischler_1981} works by iteratively selecting random subsets of the data to model inliers. It functions as an optional preprocessing step and operates in batch mode, assuming all data is available. Although RANSAC is not strictly necessary, it can enhance the performance of the standard SAM approach in certain datasets.

As we already said the core concept of SAM is to generate a counterfactual for each feature, allowing for comparison with the observed value. Anomalies are then identified based on a residual error \(\Delta_i\), which quantifies the difference between the observed and predicted values of the features:
\begin{equation}
\label{non_normalized_proximity_score}
\Delta_i = X_i - S_i
\end{equation}

or, alternatively, using a normalized version:
\begin{equation}
\label{normalized_proximity_score}
\hat \Delta_i = \frac{X_i - S_i}{S_i} 
\end{equation}

This residual error provides a quantitative measure of the deviation between a feature's actual value and its predicted value. By summing the residuals across all features, we can obtain a single value that represents the overall anomaly of the event. This approach allows for the calculation of both individual feature anomalies and an aggregate anomaly score.

For more detail on the implementation, please refer to Algorithm~\ref{alg:samfit} for the training/fitting step, and Algorithm~\ref{alg:sampredict} for inference.

\begin{algorithm}[H]
    \caption{SAM Training/Fitting Algorithm}
    \label{alg:samfit}
    \begin{algorithmic}[1]
        \Require Dataset $X$ with $d$ features, optional regressor, RANSAC configuration
        \Ensure Coefficient matrix and intercepts for anomaly detection
        \State Initialize coefficient matrix as a zero matrix of size $d \times d$
        \State Initialize intercepts as a zero vector of size $d$
        \For{each feature $i$ from $0$ to $d-1$}
            \State Select all features except $i$ as predictors
            \If{RANSAC is enabled}
                \State Initialize RANSAC regressor with specified regressor
            \Else
                \State Use provided or default regressor
            \EndIf
            \State Fit the model using predictors and target as feature $i$
            \If{Using RANSAC and it converged}
                \State Update coefficient matrix and intercepts with RANSAC model parameters
            \Else
                \State Update with ordinary regression parameters if RANSAC failed
            \EndIf
        \EndFor
        \Return Fitted model with coefficient matrix and intercepts
    \end{algorithmic}
\end{algorithm}

\begin{algorithm}[H]
    \caption{SAM Inference Algorithm}
    \label{alg:sampredict}
    \begin{algorithmic}[1]
        \Require Dataset $X$, fitted model with coefficient matrix and intercepts, threshold $\tau$
        \State Compute predicted values by transforming $X$ using the coefficient matrix and intercepts
        \State Calculate distances between predicted and actual values
        \If{Normalization is enabled}
            \State Normalize distances by dividing with actual values plus a small constant
        \EndIf
        \State Compute scores as the sum of normalized distances for each data point
        \State Determine threshold value based on the given percentile of scores
        \Return Anomaly labels: $-1$ for anomalies, $1$ for normal data, based on threshold comparison
    \end{algorithmic}
\end{algorithm}

\section{Benchmarking SAM}

We evaluated the performance of SAM by comparing it against established anomaly detection benchmark models across various datasets. We note that the competing methods: Isolation Forest, OneClass SVM, Local Outlier Factor and kNN are often pointed as the more competitive one across different datasets~\cite{bouman2024unsupervised, goldstein2016comparative}.
Since labeled data is typically not available in real-world anomaly detection tasks, we did not optimize the hyperparameters of the models. The only exception is kNN, where we set the number of neighbors to $log(n)$, where $n$ represents the number of examples in the dataset. Additionally, as we only used area metrics (such as AUC), we avoided the need to set a threshold for classification. 

\subsection{Competing Models}

The selection of competing methods was guided by a balance between well-established techniques and those demonstrated to perform optimally in numerous prior studies~\cite{goldstein2016comparative, bouman2024unsupervised}. Comprehensive analyses have highlighted the superior performance of Isolation Forest, Local Outlier Factor (LOF), and k-NN across various datasets. Additionally, One-Class SVM was included due to its continued use in applications, and its robust foundation in statistical learning theory~\cite{mello2018machine}.

The \textbf{Isolation Forest} (iForest) \cite{liu2008isolation} algorithm isolates observations by randomly selecting features and split values within their range. The fundamental idea is that anomalies, being distinct from the majority of the data, are easier to isolate than normal points.

\[
s(x, n) = 2^{-\frac{E(h(x))}{c(n)}},
\]
where \( E(h(x)) \) represents the average path length of point \( x \), \( n \) is the number of external nodes, and \( c(n) \) is the average path length of an unsuccessful search in a Binary Search Tree.

The \textbf{One-Class SVM} \cite{scholkopf2001estimating} algorithm seeks to find a decision function that distinguishes regions of high data density from regions of low density.

\[
\delta = \min_{w, \xi, \rho} \frac{1}{2} \|w\|^2 + \frac{1}{\nu n} \sum_{i=1}^n \xi_i - \rho
\]

subject to \( w \cdot \phi(x_i) \geq \rho - \xi_i \) and \( \xi_i \geq 0 \), where \( \phi(x_i) \) is the feature map induced by the kernel, \( \nu \) is an upper bound on the fraction of outliers, and \( \rho \) is the offset of the decision function.

The \textbf{Local Outlier Factor (LOF)} \cite{breunig2000lof} algorithm assesses the local density deviation of a data point relative to its neighbors.

\[
LOF_k(x) = \frac{1}{|N_k(x)|} \sum_{y \in N_k(x)} \frac{lrd_k(y)}{lrd_k(x)},
\]
where \( N_k(x) \) is the set of \( k \) nearest neighbors of \( x \), and \( lrd \) (local reachability density) is the inverse of the average reachability distance.

The \textbf{k-Nearest Neighbors (kNN)} \cite{1053964} anomaly detection algorithm works by measuring the distance between a data point and its nearest neighbors. The assumption is that normal points reside in dense regions of the feature space, whereas anomalies are isolated and distant from their neighbors. In this approach, a point's anomaly score is typically defined by the distance to its \( k \)-th nearest neighbor or by averaging the distances to its \( k \) nearest neighbors.

\[
kNN(x) = \frac{1}{k} \sum_{i=1}^{k} d(x, x_i),
\]
where \( d(x, x_i) \) represents the distance between the point \( x \) and its \( i \)-th nearest neighbor, and \( k \) is the number of nearest neighbors considered. Anomalies are identified as points with larger average distances, indicating lower density and potential isolation from the normal data distribution.

\subsection{Complexity Comparison of Anomaly Detection Methods}

We compare the computational complexity of different anomaly detection methods, namely Synthetic Anomaly Monitoring (SAM), Local Outlier Factor (LOF), k-Nearest Neighbors (KNN), and Isolation Forest, based on the number of features (\(d\)) and instances (\(n\)) to be scored online, i.e. inference time:
\begin{itemize}
    \item \textbf{Synthetic Anomaly Monitoring (SAM):} computes a dot product for each feature against the remaining features, processing the \(d\)-dimensional feature vector linearly. With pre-trained weights, inference involves linear operations. \textbf{Complexity:} \(O(d^2)\)
    
    \item \textbf{Local Outlier Factor (LOF):} computes the distance from the target instance to all other instances (\(O(n \times d)\)), and calculates local density using \(k\) nearest neighbors (\(O(k \times d)\)). \textbf{Complexity:} \(O(n \times d + k \times d)\)

    \item \textbf{k-Nearest Neighbors (KNN):} computes distances from the target instance to all \(n\) instances in \(d\)-dimensional space (\(O(n \times d)\)), and then evaluates the anomaly score based on the \(k\) nearest neighbors. Compute the score using these neighbors is \(O(k \times d)\). \textbf{Complexity:} \(O(n \times d + k \times d)\)

    \item \textbf{Isolation Forest:} requires traversal of \(t\) trees with a typical depth \(O(\log n)\), involving processing \(O(d)\) per node. The total complexity is \(O(t \cdot \log n \cdot d)\), where \(t\) is the number of trees. \textbf{Complexity:} \(O(t \cdot \log n \cdot d)\)
\end{itemize}

Therefore, LOF and KNN scale with the number of instances \(n\), making them computationally intensive in large datasets, while SAM is primarily dependent on \(d\). Isolation Forest balances instance scalability and data-dimensional sensitivity through tree-based structures.

For tabular data it is common to have a scenario of $d << n$. In online settings, SAM's linear operations in pre-trained models make it suitable for real-time anomaly detection. 

\subsection{Benchmark Datasets}

We use five real dataset to asses the performance in the real world for SAM.

\begin{enumerate}

\item \textbf{Credit Card Fraud}

This dataset is hosted on Kaggle and contains transactions made by credit cards in September 2013 by European cardholders. It presents transactions that occurred in two days, where we have 492 frauds out of 284,807 transactions. The dataset is highly unbalanced, the positive class (frauds) accounts for 0.172\% of all transactions.
 It contains only numerical input variables which are the result of a PCA transformation. Due to confidentiality issues, the original features and more background information about the data are not provided~\cite{kaggle_creditcard_fraud}.

\item \textbf{Http Attack}
This dataset is created by the Spanish Research National Council and contains thousands of web requests automatically generated to mimic normal user browsing and simulated web attacks. It’s designed to test web attack protection systems.
 The dataset contains HTTP requests in raw text, including method, URL, version, headers, and the query content~\cite{http_csic_2010}.

\item \textbf{KDD Cup 1999 Data}
 Although somewhat outdated, this dataset is a classic and still widely used in the anomaly detection community. It was used for The Third International Knowledge Discovery and Data Mining Tools Competition, which was held in conjunction with KDD-99 The Fifth International Conference on Knowledge Discovery and Data Mining. The task was to build a network intrusion detector, a predictive model capable of distinguishing between "bad" connections, called intrusions or attacks, and "good" normal connections.
It includes a wide variety of intrusions simulated in a military network environment. Each connection is labeled as either normal or an attack, with exactly one specific attack type. The dataset contains about 41 features\cite{kdd_cup_1999}.

\item \textbf{Forest Cover}
This dataset predicts forest cover type using cartographic variables, excluding remotely sensed data. The cover type for each 30x30 meter observation cell was determined using USFS Region 2 data, while the independent variables were derived from USGS and USFS sources. The data is in raw form, with binary columns for qualitative variables (e.g., wilderness areas, soil types). The study area includes four wilderness regions in Roosevelt National Forest, Colorado, each with distinct elevation ranges and tree species, reflecting minimal human disturbances and natural ecological processes. \cite{rt7n-2x60-20, covertype_31}

\item \textbf{MulCross}
Mulcross dataset, which can be downloaded in CSV format from OpenML \cite{mulcross} is a synthetic dataset generated using a multivariate normal distribution and contains a sufficient number of anomalies. It includes 262,144 data points, with 10 percent classified as anomalies. Each data point is composed of four different features. \cite{rt7n-2x60-20}

\end{enumerate}

\subsection{Experimental Setup}

To evaluate the performance of the proposed model and competing methods, for each experimental run, the following procedure was conducted:
\begin{enumerate}
    \item \textbf{Bootstrapping} of the original dataset;
    \item \textbf{Train-Test Splitting}: each bootstrapped sample was randomly split into training and testing subsets, with 70\% of the data utilized for training/validation the model, while the remaining 30\% was retained for testing;
    \item \textbf{Repetition}: to account for variability and to enhance the statistical rigor of our findings, this process of bootstrapping and train-test splitting was repeated 10 times for each experiment;
    \item \textbf{Performance Evaluation}: The model's performance was evaluated using ROC AUC and PR AUC metrics averaged over the 15 repetitions, ensured that the results were not biased by any single train-test split configuration.
\end{enumerate}

By utilizing bootstrapping, coupled with repeated random train-test splits, we aimed to obtain a robust and unbiased evaluation of the model's efficacy in anomaly detection tasks. This systematic approach not only facilitates a thorough validation of results but also enriches the credibility and generalizability of the findings.

Our choice of ROC AUC and PR AUC metrics was to assess how effectively the models distinguish between normal and anomalous instances, while handling imbalanced data without requiring a fixed threshold.

For the distance-based methods $k$ was set to be $\log{n}$, where $n$ is the dataset size.

We employed four versions of SAM:
\begin{itemize}
    \item \textbf{SAM$^{++}$}: uses RANSAC and the normalized proximity score (see Equation \ref{normalized_proximity_score}).
    \item \textbf{SAM$^{+-}$}: uses RANSAC with the non-normalized proximity score (see Equation \ref{non_normalized_proximity_score}).
    \item \textbf{SAM$^{-+}$}: does not use RANSAC, and employs the normalized proximity score (see Equation \ref{normalized_proximity_score}).
    \item \textbf{SAM$^{--}$}: does not use RANSAC nor the normalized proximity score, opting instead for the non-normalized proximity score (see Equation \ref{non_normalized_proximity_score}).
\end{itemize}

\begin{figure}[h!]
 \label{figure1}
    \centering
    % Adjust the width of the framebox as needed
    %\framebox{
        \begin{minipage}[c]{0.8\linewidth}
            \centering
            \includegraphics[width=\linewidth]{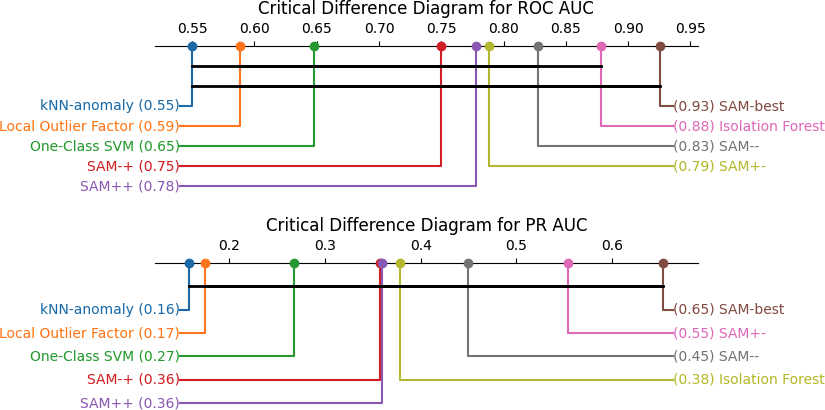}
            \caption{Critical difference plots for each method across all datasets, considering ROC AUC and PR AUC metrics. SAM-best considers the best scenario of SAM for each dataset ($p$-value$=0.07$).}
            \label{fig:CDplot}
        \end{minipage}
    %}
\end{figure}

\begin{figure}[ht!]
 \label{figure1}
    \centering
    % Adjust the width of the framebox as needed
    %\framebox{
        \begin{minipage}[c]{1.1\linewidth}
            \centering
            \includegraphics[width=\linewidth]{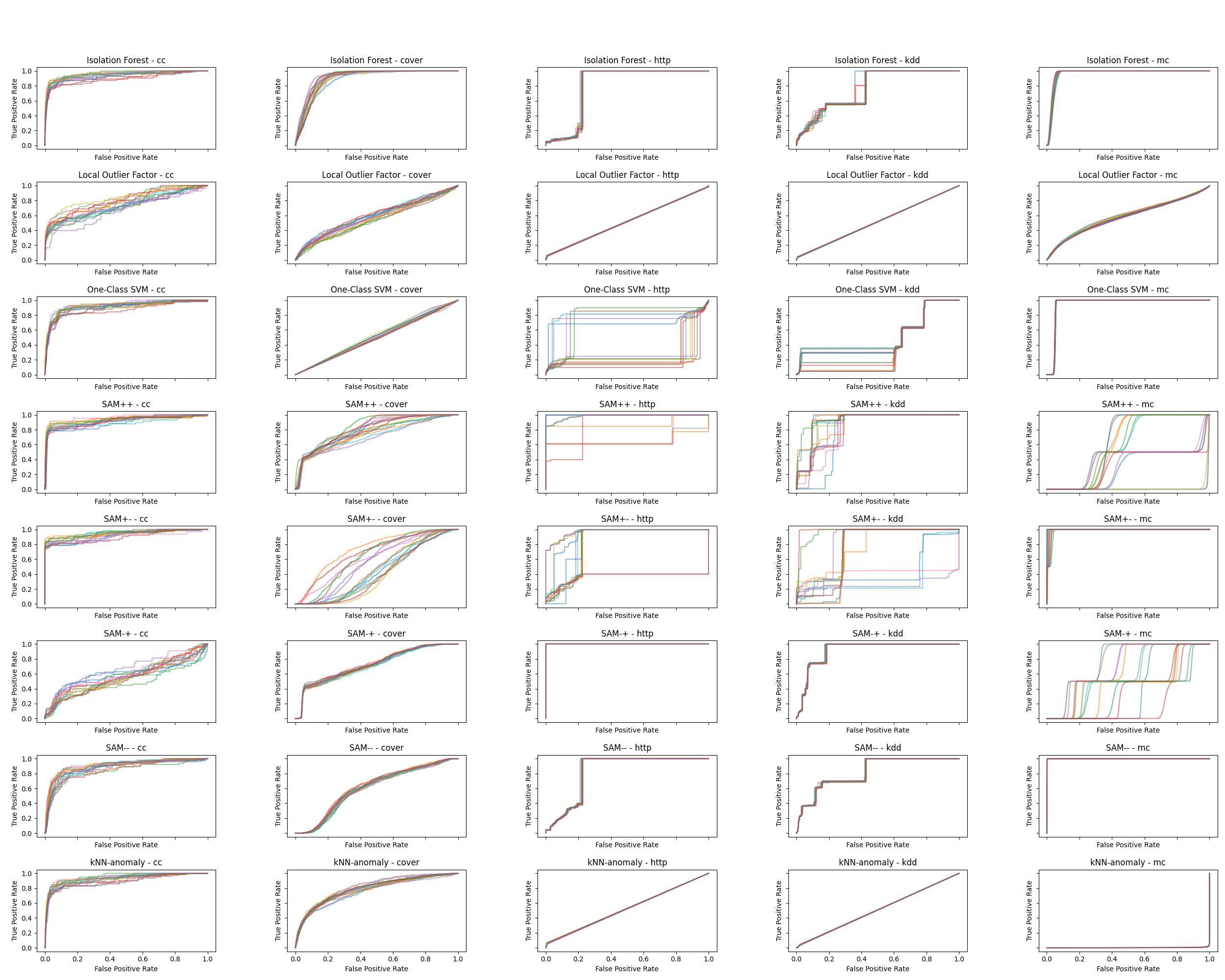}
            \caption{ROC AUC Curves of the experiment}
            \label{fig:rocauc}            
        \end{minipage}
    %}
\end{figure}

\begin{figure}[ht!]
    \label{figure2}
    \centering
    % Adjust the width of the frame box as needed
    %\framebox{
        \begin{minipage}[c]{1.2\linewidth}
            \centering
            \includegraphics[width=\linewidth]{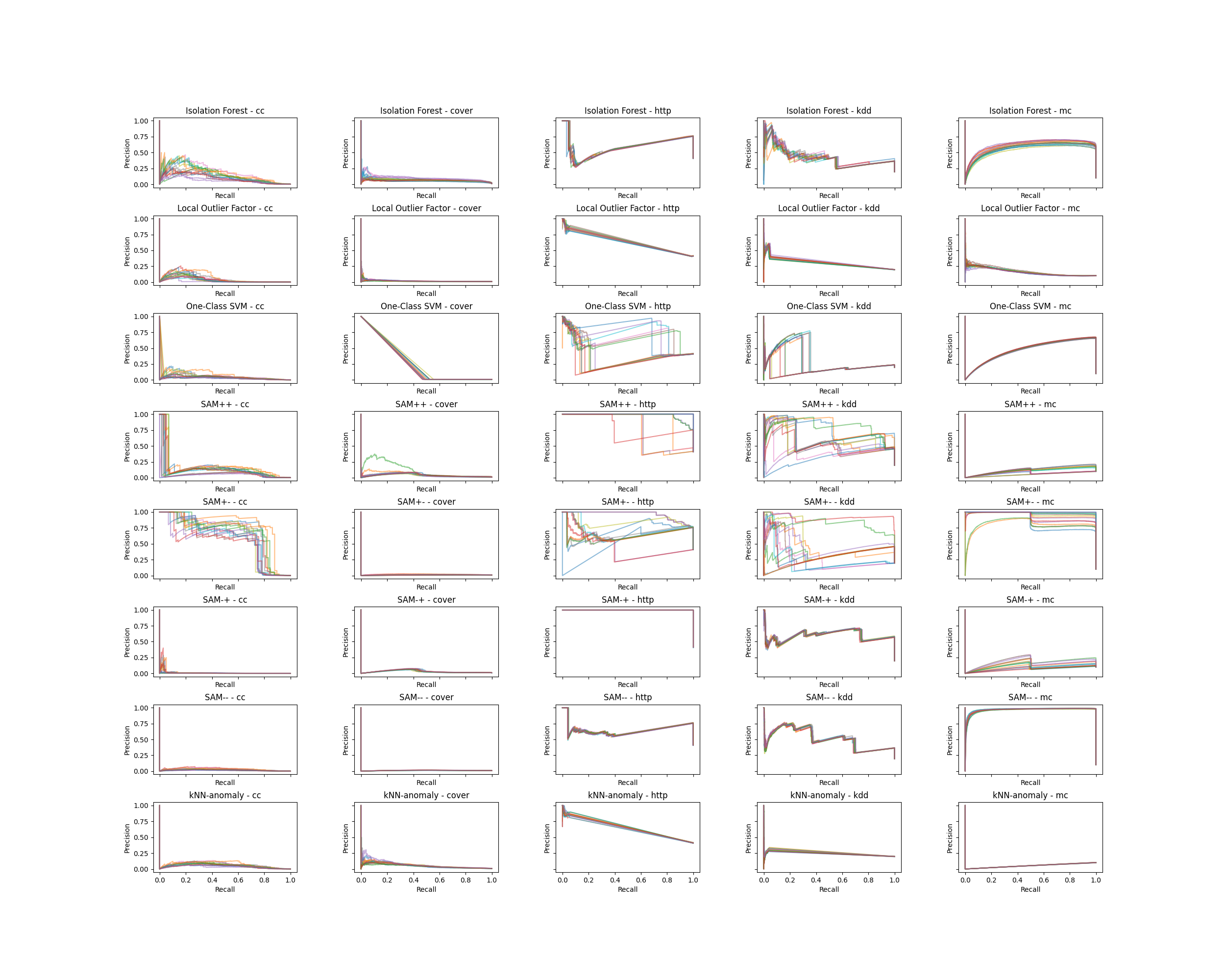}
            \caption{PR AUC Curves of the experiment}
            \label{fig:prauc}
        \end{minipage}
    %}
\end{figure}

 The results, summarized in Tables \ref{tab:roc_auc_comparison} and \ref{tab:pr_auc_comparison}, provide insights into how these factors affect the SAM model's ability to detect anomalies. All the ROC and PR curves are available in Figures~\ref{fig:rocauc} and ~\ref{fig:prauc}. Additionally, critical difference plots are shown in Figure~\ref{fig:CDplot} for both ROC AUC and PRAUC metrics.
 
\subsubsection{Effect of RANSAC and Normalization}

The analysis of the SAM variants highlights the critical roles of both RANSAC and proximity score normalization in enhancing anomaly detection performance. In particular, SAM$^{++}$ (RANSAC + normalized proximity score) and SAM$^{-+}$ (no RANSAC + normalized proximity score) demonstrated consistently high ROC AUC and PR AUC scores on datasets like "http" and "kdd." This suggests that normalization effectively stabilizes the detection process by accounting for varying feature scales, which is often crucial in datasets with diverse characteristics.

\textbf{RANSAC's Impact}: The use of RANSAC appears to enhance the model's robustness, particularly in handling noisy data. For example, SAM$^{++}$ achieved high ROC AUC scores on "cc" (0.92 $\pm$ 0.04) and "http" (0.91 $\pm$ 0.30), indicating that RANSAC’s outlier filtering helps to identify anomalies more effectively in certain contexts. However, the variability in performance, as seen in the large standard deviations (e.g., SAM$^{++}$ on "http"), suggests that RANSAC's stochastic nature can lead to inconsistent results depending on the dataset's complexity and noise levels.

\textbf{Normalization’s Role}: Normalizing the proximity scores had a noticeable effect on performance, especially in datasets with varying feature scales. SAM variants with normalization (SAM$^{++}$ and SAM$^{-+}$) outperformed their non-normalized counterparts (SAM$^{+-}$ and SAM$^{--}$) in most datasets. For instance, SAM$^{-+}$ achieved a perfect PR AUC score on "http" (1.00 $\pm$ 0.00) and a high ROC AUC on "kdd" (0.92 $\pm$ 0.00). This suggests that normalization not only mitigates the influence of feature scaling but may enhance the model's capability to distinguish between normal and anomalous data points, especially in high-dimensional spaces.

\subsubsection{Performance Variability Across Datasets}

The performance of SAM variants varied considerably across different datasets, reflecting the influence of data characteristics such as dimensionality, noise levels, and class imbalance. In datasets with clear anomaly structures, such as "http" and "kdd," the SAM variants, particularly SAM$^{++}$ and SAM$^{-+}$, achieved high scores. This suggests that normalization and robust regression (RANSAC) are effective in contexts where anomalies are well-separated from normal data points.

Conversely, in more complex or noisy datasets like "cc" and "cover," the performance was less consistent. SAM$^{+-}$ (RANSAC + non-normalized) exhibited high scores on "mc" (ROC AUC: 0.99 $\pm$ 0.01) but struggled on "cover" (ROC AUC: 0.53 $\pm$ 0.20). The elevated standard deviations observed in some cases (e.g., SAM$^{+-}$ on "http") indicate that certain combinations of RANSAC and normalization may lead to overfitting or sensitivity to particular data distributions.

While this study provides valuable insights into the performance of different SAM variants, the variability in results highlights the need for additional research. Future work could explore the impact of different RANSAC configurations and proximity normalization techniques. Moreover, expanding the analysis to include a broader range of datasets with varying characteristics would help generalize the findings and refine model selection guidelines.

In summary, the analysis underscores the importance of dataset characteristics in choosing the optimal SAM variant. RANSAC and normalization significantly influence the model's performance, with SAM$^{++}$ and SAM$^{-+}$ showing the most promise for robust anomaly detection in high-dimensional and diverse datasets. Although no statistical difference could be found when considered a Friedman rank-based test  ($p$-value$=0.07$), we can see from Figure~\ref{fig:CDplot} that SAM often appears high in the ranks and is competitive with the de-facto state-of-the-art method Isolation Forest~\cite{bouman2024unsupervised}.

% Overall, SAM exhibited mixed results depending on the implementation details and the specific datasets. The versions using RANSAC, whether with normalized or non-normalized proximity scores, performed well on the credit fraud dataset and consistently showed strong performance on the others, though they never stood out as the top-performing model.

% In contrast, the versions without RANSAC achieved perfect scores on two of the five datasets, highlighting their potential in certain scenarios. While other models, such as Isolation Forest, delivered impressive results across all datasets, none managed to achieve a perfect score in any of them. This variation in performance across different datasets and metrics underscores SAM's versatility in addressing a wide range of data challenges, provided that the appropriate version is selected.

\clearpage
\begin{table}[h]
    \centering
    
    \begin{tabular}{llllll} \toprule datasets & cc & cover & http & kdd & mc \\   &   &   &   &   &   \\ Model &  &  &  &  &  \\ \midrule Isolation Forest & \textbf{0.94} ± 0.05 & \textbf{0.91} ± 0.03 & 0.80 ± 0.01 & 0.77 ± 0.02 & 0.97 ± 0.01 \\ 
    Local Outlier Factor & 0.75 ± 0.09 & 0.57 ± 0.05 & 0.52 ± 0.01 & 0.51 ± 0.00 & 0.59 ± 0.02 \\ 
    One-Class SVM & 0.91 ± 0.04 & 0.50 ± 0.03 & 0.45 ± 0.50 & 0.43 ± 0.10 & 0.95 ± 0.00 \\ 
    SAM$^{++}$ & 0.92 ± 0.04 & 0.78 ± 0.08 & \underline{0.91} ± 0.30 & \underline{0.88} ± 0.10 & 0.40 ± 0.50 \\ 
    SAM$^{+-}$ & \underline{0.93} ± 0.04 & 0.53 ± 0.2 & 0.80 ± 0.40 & 0.69 ± 0.40 & \underline{0.99} ± 0.01 \\ 
    SAM$^{-+}$ & 0.56 ± 0.09 & 0.73 ± 0.02 & \textbf{1.00} ± 0.00 & \textbf{0.92} ± 0.00 & 0.54 ± 0.30 \\ 
    SAM$^{--}$ & 0.89 ± 0.05 & 0.60 ± 0.02 & 0.83 ± 0.01 & 0.82 ± 0.01 & \textbf{1.00} ± 0.00 \\ 
    kNN-anomaly & 0.92 ± 0.04 & \underline{0.79} ± 0.03 & 0.53 ± 0.01 & 0.51 ± 0.00 & 0.00 ± 0.00 \\ \bottomrule \end{tabular}
    \caption{ROC AUC Comparison Across Datasets. SAM$^{++}$ using RANSAC and normalized proximity score \ref{normalized_proximity_score}, SAM$^{+-}$ using RANSAC and non normalized proximity score \ref{non_normalized_proximity_score}. SAM$^{-+}$ using no RANSAC and normalized proximity score. SAM$^{--}$ no RANSAC and no normalized proximity score. The values represent mean ± $2\times$ standard deviation. Bold values indicate the best, and underlined values the second best approaches.}
    \label{tab:roc_auc_comparison}
\end{table}

\begin{table}[h]
    \centering
    
    \begin{tabular}{llllll} \toprule datasets & cc & cover & http & kdd & mc \\   &   &   &   &   &   \\ Model &  &  &  &  &  \\ \midrule 
    Isolation Forest & \underline{0.13} ± 0.09 & \textbf{0.07} ± 0.03 & 0.66 ± 0.01 & 0.45 ± 0.03 & 0.58 ± 0.06 \\ 
    Local Outlier Factor & 0.04 ± 0.03 & 0.02 ± 0.01 & 0.44 ± 0.010 & 0.21 ± 0.01 & 0.16 ± 0.01 \\ 
    One-Class SVM & 0.04 ± 0.03 & 0.01 ± 0.00 & 0.57 ± 0.30 & 0.25 ± 0.1 & 0.47 ± 0.01 \\ 
    SAM$^{++}$ & 0.12 ± 0.08 & 0.04 ± 0.04 & \underline{0.93} ± 0.20 & \underline{0.62} ± 0.30 & 0.09 ± 0.06 \\ 
    SAM$^{+-}$ & \textbf{0.63} ± 0.20 & 0.01 ± 0.01 & 0.75 ± 0.30 & 0.44 ± 0.4 & \underline{0.94} ± 0.10 \\ 
    SAM$^{-+}$ & 0.01 ± 0.01 & 0.03 ± 0.01 & \textbf{1.00} ± 0.00 & \textbf{0.63} ± 0.01 & 0.12 ± 0.08 \\ 
    SAM$^{--}$ & 0.02 ± 0.02 & 0.01 ± 0.00 & 0.72 ± 0.01 & 0.53 ± 0.01 & \textbf{0.97} ± 0.01 \\ 
    kNN-anomaly & 0.05 ± 0.03 & \underline{0.05} ± 0.02 & 0.44 ± 0.01 & 0.20 ± 0.00 & 0.05 ± 0.00 \\ \bottomrule \end{tabular} 
    \caption{PR AUC Comparison Across Datasets.  SAM$^{++}$ using RANSAC and normalized proximity score \ref{normalized_proximity_score}, SAM$^{+-}$ using RANSAC and non normalized proximity score \ref{non_normalized_proximity_score}. SAM$^{-+}$ using no RANSAC and normalized proximity score. SAM$^{--}$ no RANSAC and no normalized proximity score. The values represent mean ± $2\times$ standard deviation. Bold values indicate the best, and underlined values the second best approaches. }
    \label{tab:pr_auc_comparison}
\end{table}

%\clearpage
\section{Conclusion}

In this paper, we introduced and evaluated Synthetic Anomaly Monitoring (SAM), a novel approach to anomaly detection that leverages synthetic control methods from causal inference. We benchmarked SAM against well-established models, including Isolation Forest, Local Outlier Factor (LOF), k-NN, and One-Class SVM, using five widely-used datasets: Credit Card Fraud (cc), the HTTP Dataset CSIC 2010 (http), Forest Cover (cover), Mulcross (mulcross), and KDD Cup 1999 (kdd).

SAM demonstrated strong performance, outperforming other models in terms of ROC AUC on the \textit{http}, \textit{kdd}, and \textit{mulcross} datasets. This indicates SAM's ability to accurately capture the underlying data structure and effectively detect anomalies. Even more, SAM showed excellent results in PR AUC evaluations, excelling in four of the five datasets, which highlights its effectiveness in handling imbalanced datasets.

Overall, SAM's consistent performance underscores its potential as a robust tool for anomaly detection, providing a reliable alternative to traditional methods. Its capability to maintain high detection accuracy across different datasets and varying noise levels makes it particularly well-suited for real-world applications, where data complexity and imbalance are frequent challenges. Future research will focus on extending SAM to other data types and integrating it into real-time monitoring systems.

While other methods may excel in specific scenarios, SAM's consistency across various challenging environments establishes it as a dependable choice for general-purpose anomaly detection tasks. In sum, SAM stands out as a versatile and pragmatic approach, ideal for scenarios demanding steady performance across an array of conditions.

As a first study in synthetic anomaly monitoring, we focused primarily on outlining the methodology and evaluating its effectiveness. Nevertheless, there remain avenues for future exploration. In particular, the model's explainability, online learning and incremental adaptation capabilities. These dimensions promise to fortify SAM's practicality and deserve further investigation.

\backmatter

\bmhead{Acknowledgements}

The authors would like to express their gratitude to Pedro Carossi for his invaluable support and to Mercadolibre for providing the opportunity and resources necessary to conduct this investigation. We also extend our thanks to all the collaborators at Meliminds for their insightful comments and feedback throughout the research process.

%%\bibliography{sn-bibliography}% common bib file

\begin{thebibliography}{27}
% BibTex style file: bmc-mathphys.bst (version 2.1), 2014-07-24
\ifx \bisbn   \undefined \def \bisbn  #1{ISBN #1}\fi
\ifx \binits  \undefined \def \binits#1{#1}\fi
\ifx \bauthor  \undefined \def \bauthor#1{#1}\fi
\ifx \batitle  \undefined \def \batitle#1{#1}\fi
\ifx \bjtitle  \undefined \def \bjtitle#1{#1}\fi
\ifx \bvolume  \undefined \def \bvolume#1{\textbf{#1}}\fi
\ifx \byear  \undefined \def \byear#1{#1}\fi
\ifx \bissue  \undefined \def \bissue#1{#1}\fi
\ifx \bfpage  \undefined \def \bfpage#1{#1}\fi
\ifx \blpage  \undefined \def \blpage #1{#1}\fi
\ifx \burl  \undefined \def \burl#1{\textsf{#1}}\fi
\ifx \doiurl  \undefined \def \doiurl#1{\url{https://doi.org/#1}}\fi
\ifx \betal  \undefined \def \betal{\textit{et al.}}\fi
\ifx \binstitute  \undefined \def \binstitute#1{#1}\fi
\ifx \binstitutionaled  \undefined \def \binstitutionaled#1{#1}\fi
\ifx \bctitle  \undefined \def \bctitle#1{#1}\fi
\ifx \beditor  \undefined \def \beditor#1{#1}\fi
\ifx \bpublisher  \undefined \def \bpublisher#1{#1}\fi
\ifx \bbtitle  \undefined \def \bbtitle#1{#1}\fi
\ifx \bedition  \undefined \def \bedition#1{#1}\fi
\ifx \bseriesno  \undefined \def \bseriesno#1{#1}\fi
\ifx \blocation  \undefined \def \blocation#1{#1}\fi
\ifx \bsertitle  \undefined \def \bsertitle#1{#1}\fi
\ifx \bsnm \undefined \def \bsnm#1{#1}\fi
\ifx \bsuffix \undefined \def \bsuffix#1{#1}\fi
\ifx \bparticle \undefined \def \bparticle#1{#1}\fi
\ifx \barticle \undefined \def \barticle#1{#1}\fi
\bibcommenthead
\ifx \bconfdate \undefined \def \bconfdate #1{#1}\fi
\ifx \botherref \undefined \def \botherref #1{#1}\fi
\ifx \url \undefined \def \url#1{\textsf{#1}}\fi
\ifx \bchapter \undefined \def \bchapter#1{#1}\fi
\ifx \bbook \undefined \def \bbook#1{#1}\fi
\ifx \bcomment \undefined \def \bcomment#1{#1}\fi
\ifx \oauthor \undefined \def \oauthor#1{#1}\fi
\ifx \citeauthoryear \undefined \def \citeauthoryear#1{#1}\fi
\ifx \endbibitem  \undefined \def \endbibitem {}\fi
\ifx \bconflocation  \undefined \def \bconflocation#1{#1}\fi
\ifx \arxivurl  \undefined \def \arxivurl#1{\textsf{#1}}\fi
\csname PreBibitemsHook\endcsname

%%% 1
\bibitem[\protect\citeauthoryear{Hawkins and Hawkins}{1980}]{Hawkins_1980}
\begin{barticle}
\bauthor{\bsnm{Hawkins}, \binits{D.M.}},
\bauthor{\bsnm{Hawkins}, \binits{D.M.}}:
\batitle{Identification of outliers}.
\bjtitle{null}
(\byear{1980})
\doiurl{null}
\end{barticle}
\endbibitem

%%% 2
\bibitem[\protect\citeauthoryear{Goldstein and
  Uchida}{2016}]{goldstein2016comparative}
\begin{barticle}
\bauthor{\bsnm{Goldstein}, \binits{M.}},
\bauthor{\bsnm{Uchida}, \binits{S.}}:
\batitle{A comparative evaluation of unsupervised anomaly detection algorithms
  for multivariate data}.
\bjtitle{PloS one}
\bvolume{11}(\bissue{4}),
\bfpage{0152173}
(\byear{2016})
\end{barticle}
\endbibitem

%%% 3
\bibitem[\protect\citeauthoryear{Schmidl et~al.}{2022}]{Schmidl_2022}
\begin{barticle}
\bauthor{\bsnm{Schmidl}, \binits{S.}},
\bauthor{\bsnm{Schmidl}, \binits{S.}},
\bauthor{\bsnm{Wenig}, \binits{P.}},
\bauthor{\bsnm{Wenig}, \binits{P.}},
\bauthor{\bsnm{Papenbrock}, \binits{T.}},
\bauthor{\bsnm{Papenbrock}, \binits{T.}}:
\batitle{Anomaly detection in time series}.
\bjtitle{Proceedings of the VLDB Endowment}
(\byear{2022})
\doiurl{10.14778/3538598.3538602}
\end{barticle}
\endbibitem

%%% 4
\bibitem[\protect\citeauthoryear{Chandola
  et~al.}{2009}]{10.1145/1541880.1541882}
\begin{botherref}
\oauthor{\bsnm{Chandola}, \binits{V.}},
\oauthor{\bsnm{Banerjee}, \binits{A.}},
\oauthor{\bsnm{Kumar}, \binits{V.}}:
Anomaly detection: A survey.
Association for Computing Machinery
\textbf{41}(3)
(2009)
\doiurl{10.1145/1541880.1541882}
\end{botherref}
\endbibitem

%%% 5
\bibitem[\protect\citeauthoryear{Calikus et~al.}{2022}]{calikus2022wisdom}
\begin{barticle}
\bauthor{\bsnm{Calikus}, \binits{E.}},
\bauthor{\bsnm{Nowaczyk}, \binits{S.}},
\bauthor{\bsnm{Bouguelia}, \binits{M.-R.}},
\bauthor{\bsnm{Dikmen}, \binits{O.}}:
\batitle{Wisdom of the contexts: active ensemble learning for contextual
  anomaly detection}.
\bjtitle{Data Mining and Knowledge Discovery}
\bvolume{36}(\bissue{6}),
\bfpage{2410}--\blpage{2458}
(\byear{2022})
\end{barticle}
\endbibitem

%%% 6
\bibitem[\protect\citeauthoryear{Li and Van~Leeuwen}{2023}]{li2023explainable}
\begin{barticle}
\bauthor{\bsnm{Li}, \binits{Z.}},
\bauthor{\bsnm{Van~Leeuwen}, \binits{M.}}:
\batitle{Explainable contextual anomaly detection using quantile regression
  forests}.
\bjtitle{Data Mining and Knowledge Discovery}
\bvolume{37}(\bissue{6}),
\bfpage{2517}--\blpage{2563}
(\byear{2023})
\end{barticle}
\endbibitem

%%% 7
\bibitem[\protect\citeauthoryear{Steinbuss and
  B{\"o}hm}{2021}]{steinbuss2021benchmarking}
\begin{barticle}
\bauthor{\bsnm{Steinbuss}, \binits{G.}},
\bauthor{\bsnm{B{\"o}hm}, \binits{K.}}:
\batitle{Benchmarking unsupervised outlier detection with realistic synthetic
  data}.
\bjtitle{ACM Transactions on Knowledge Discovery from Data (TKDD)}
\bvolume{15}(\bissue{4}),
\bfpage{1}--\blpage{20}
(\byear{2021})
\end{barticle}
\endbibitem

%%% 8
\bibitem[\protect\citeauthoryear{Bouman et~al.}{2024}]{bouman2024unsupervised}
\begin{barticle}
\bauthor{\bsnm{Bouman}, \binits{R.}},
\bauthor{\bsnm{Bukhsh}, \binits{Z.}},
\bauthor{\bsnm{Heskes}, \binits{T.}}:
\batitle{Unsupervised anomaly detection algorithms on real-world data: how many
  do we need?}
\bjtitle{Journal of Machine Learning Research}
\bvolume{25}(\bissue{105}),
\bfpage{1}--\blpage{34}
(\byear{2024})
\end{barticle}
\endbibitem

%%% 9
\bibitem[\protect\citeauthoryear{Costa et~al.}{2013}]{costa2013partially}
\begin{bchapter}
\bauthor{\bsnm{Costa}, \binits{G.B.}},
\bauthor{\bsnm{Ponti}, \binits{M.}},
\bauthor{\bsnm{Frery}, \binits{A.C.}}:
\bctitle{Partially supervised anomaly detection using convex hulls on a 2d
  parameter space}.
In: \bbtitle{Partially Supervised Learning: Second IAPR International Workshop,
  PSL 2013, Nanjing, China, May 13-14, 2013, Revised Selected Papers 2},
pp. \bfpage{1}--\blpage{8}
(\byear{2013}).
\bcomment{Springer}
\end{bchapter}
\endbibitem

%%% 10
\bibitem[\protect\citeauthoryear{Sch{\"o}lkopf
  et~al.}{2001}]{scholkopf2001estimating}
\begin{barticle}
\bauthor{\bsnm{Sch{\"o}lkopf}, \binits{B.}},
\bauthor{\bsnm{Williamson}, \binits{R.C.}},
\bauthor{\bsnm{Bartlett}, \binits{P.L.}}:
\batitle{Estimating the support of a probability distribution}.
\bjtitle{IEEE transactions on information theory}
\bvolume{47}(\bissue{5}),
\bfpage{2531}--\blpage{2542}
(\byear{2001})
\end{barticle}
\endbibitem

%%% 11
\bibitem[\protect\citeauthoryear{Tax and Duin}{2004}]{tax2004support}
\begin{barticle}
\bauthor{\bsnm{Tax}, \binits{D.M.}},
\bauthor{\bsnm{Duin}, \binits{R.P.}}:
\batitle{Support vector data description}.
\bjtitle{Machine learning}
\bvolume{54},
\bfpage{45}--\blpage{66}
(\byear{2004})
\end{barticle}
\endbibitem

%%% 12
\bibitem[\protect\citeauthoryear{Liu et~al.}{2008}]{liu2008isolation}
\begin{bchapter}
\bauthor{\bsnm{Liu}, \binits{F.-T.}},
\bauthor{\bsnm{Ting}, \binits{K.M.}},
\bauthor{\bsnm{Zhou}, \binits{Z.-H.}}:
\bctitle{Isolation forest}.
In: \bbtitle{Proceedings of the 2008 8th IEEE International Conference on Data
  Mining (ICDM)},
pp. \bfpage{413}--\blpage{422}
(\byear{2008}).
\bcomment{IEEE}
\end{bchapter}
\endbibitem

%%% 13
\bibitem[\protect\citeauthoryear{Ramaswamy
  et~al.}{2000}]{ramaswamy2000efficient}
\begin{bchapter}
\bauthor{\bsnm{Ramaswamy}, \binits{S.}},
\bauthor{\bsnm{Rastogi}, \binits{R.}},
\bauthor{\bsnm{Shim}, \binits{K.}}:
\bctitle{Efficient algorithms for mining outliers from large data sets}.
In: \bbtitle{Proceedings of the 2000 ACM SIGMOD International Conference on
  Management of Data},
pp. \bfpage{427}--\blpage{438}
(\byear{2000})
\end{bchapter}
\endbibitem

%%% 14
\bibitem[\protect\citeauthoryear{Abadie}{2021}]{10.1257/jel.20191450}
\begin{barticle}
\bauthor{\bsnm{Abadie}, \binits{A.}}:
\batitle{Using synthetic controls: Feasibility, data requirements, and
  methodological aspects}.
\bjtitle{Journal of Economic Literature}
\bvolume{59}(\bissue{2}),
\bfpage{391}--\blpage{425}
(\byear{2021})
\doiurl{10.1257/jel.20191450}
\end{barticle}
\endbibitem

%%% 15
\bibitem[\protect\citeauthoryear{Abadie
  et~al.}{2007}]{10.1198/jasa.2009.ap08746}
\begin{barticle}
\bauthor{\bsnm{Abadie}, \binits{A.}},
\bauthor{\bsnm{Diamond}, \binits{A.}},
\bauthor{\bsnm{Hainmueller}, \binits{J.}}:
\batitle{Synthetic control methods for comparative case studies: Estimating the
  effect of california's tobacco control program}.
\bjtitle{Journal of the American Statistical Association}
\bvolume{105},
\bfpage{493}--\blpage{505}
(\byear{2007})
\doiurl{10.1198/jasa.2009.ap08746}
\end{barticle}
\endbibitem

%%% 16
\bibitem[\protect\citeauthoryear{Pereira et~al.}{2020}]{pereira2020reviewing}
\begin{barticle}
\bauthor{\bsnm{Pereira}, \binits{R.C.}},
\bauthor{\bsnm{Santos}, \binits{M.S.}},
\bauthor{\bsnm{Rodrigues}, \binits{P.P.}},
\bauthor{\bsnm{Abreu}, \binits{P.H.}}:
\batitle{Reviewing autoencoders for missing data imputation: Technical trends,
  applications and outcomes}.
\bjtitle{Journal of Artificial Intelligence Research}
\bvolume{69},
\bfpage{1255}--\blpage{1285}
(\byear{2020})
\end{barticle}
\endbibitem

%%% 17
\bibitem[\protect\citeauthoryear{Resende and Ponti}{2022}]{resende2022robust}
\begin{barticle}
\bauthor{\bsnm{Resende}, \binits{D.C.O.d.}},
\bauthor{\bsnm{Ponti}, \binits{M.A.}}:
\batitle{Robust image features for classification and zero-shot tasks by
  merging visual and semantic attributes}.
\bjtitle{Neural Computing and Applications}
\bvolume{34}(\bissue{6}),
\bfpage{4459}--\blpage{4471}
(\byear{2022})
\end{barticle}
\endbibitem

%%% 18
\bibitem[\protect\citeauthoryear{Fischler et~al.}{1981}]{Fischler_1981}
\begin{barticle}
\bauthor{\bsnm{Fischler}, \binits{M.A.}},
\bauthor{\bsnm{Fischler}, \binits{M.A.}},
\bauthor{\bsnm{Fischler}, \binits{M.A.}},
\bauthor{\bsnm{Bolles}, \binits{R.C.}},
\bauthor{\bsnm{Bolles}, \binits{R.C.}}:
\batitle{Random sample consensus: a paradigm for model fitting with
  applications to image analysis and automated cartography}.
\bjtitle{Communications of The ACM}
(\byear{1981})
\doiurl{10.1145/358669.358692}
\end{barticle}
\endbibitem

%%% 19
\bibitem[\protect\citeauthoryear{Mello and Ponti}{2018}]{mello2018machine}
\begin{bbook}
\bauthor{\bsnm{Mello}, \binits{R.F.}},
\bauthor{\bsnm{Ponti}, \binits{M.A.}}:
\bbtitle{Machine Learning: a Practical Approach on the Statistical Learning
  Theory}.
\bpublisher{Springer}, \blocation{???}
(\byear{2018})
\end{bbook}
\endbibitem

%%% 20
\bibitem[\protect\citeauthoryear{Breunig et~al.}{2000}]{breunig2000lof}
\begin{bchapter}
\bauthor{\bsnm{Breunig}, \binits{M.M.}},
\bauthor{\bsnm{Kriegel}, \binits{H.-P.}},
\bauthor{\bsnm{Ng}, \binits{R.T.}},
\bauthor{\bsnm{Zimek}, \binits{J.}}:
\bctitle{Lof: Clustering-based outlier detection}.
In: \bbtitle{Proceedings of the 2000 ACM SIGMOD International Conference on
  Management of Data},
pp. \bfpage{93}--\blpage{104}
(\byear{2000})
\end{bchapter}
\endbibitem

%%% 21
\bibitem[\protect\citeauthoryear{Cover and Hart}{1967}]{1053964}
\begin{barticle}
\bauthor{\bsnm{Cover}, \binits{T.}},
\bauthor{\bsnm{Hart}, \binits{P.}}:
\batitle{Nearest neighbor pattern classification}.
\bjtitle{IEEE Transactions on Information Theory}
\bvolume{13}(\bissue{1}),
\bfpage{21}--\blpage{27}
(\byear{1967})
\doiurl{10.1109/TIT.1967.1053964}
\end{barticle}
\endbibitem

%%% 22
\bibitem[\protect\citeauthoryear{Pozzolo}{2015}]{kaggle_creditcard_fraud}
\begin{botherref}
\oauthor{\bsnm{Pozzolo}, \binits{A.D.}}:
Credit Card Fraud Detection.
\url{https://www.kaggle.com/datasets/mlg-ulb/creditcardfraud}
(2015).
\url{https://www.kaggle.com/datasets/mlg-ulb/creditcardfraud}
\end{botherref}
\endbibitem

%%% 23
\bibitem[\protect\citeauthoryear{de~Tecnologías de~la
  Comunicación~(INTECO)}{2010}]{http_csic_2010}
\begin{botherref}
\oauthor{\bsnm{Comunicación~(INTECO)}, \binits{I.N.}}:
HTTP Dataset CSIC 2010.
Accessed: 2024-08-23
(2010).
\url{https://www.isi.csic.es/dataset/}
\end{botherref}
\endbibitem

%%% 24
\bibitem[\protect\citeauthoryear{Cup}{1999}]{kdd_cup_1999}
\begin{botherref}
\oauthor{\bsnm{Cup}, \binits{K.}}:
KDD Cup 1999 Data
(1999).
\url{http://kdd.ics.uci.edu/databases/kddcup99/kddcup99.html}
\end{botherref}
\endbibitem

%%% 25
\bibitem[\protect\citeauthoryear{Jain and Jain}{2020}]{rt7n-2x60-20}
\begin{botherref}
\oauthor{\bsnm{Jain}, \binits{P.}},
\oauthor{\bsnm{Jain}, \binits{S.}}:
Anomaly detection dataset.
IEEE Dataport
(2020).
\doiurl{10.21227/rt7n-2x60} .
\url{https://dx.doi.org/10.21227/rt7n-2x60}
\end{botherref}
\endbibitem

%%% 26
\bibitem[\protect\citeauthoryear{Blackard}{1998}]{covertype_31}
\begin{botherref}
\oauthor{\bsnm{Blackard}, \binits{J.}}:
{Covertype}.
UCI Machine Learning Repository.
{DOI}: https://doi.org/10.24432/C50K5N
(1998)
\end{botherref}
\endbibitem

%%% 27
\bibitem[\protect\citeauthoryear{}{}]{mulcross}
\begin{botherref}
Mulcross.
https://www.openml.org/d/40897
\end{botherref}
\endbibitem

\end{thebibliography}
%% if required, the content of .bbl file can be included here once bbl is generated
%% BioMed_Central_Bib_Style_v1.01

\end{document}